%% file: collas2022_conference.tex
\documentclass{article} 
\usepackage{collas2022_conference,times}

\input{math_commands.tex}

\usepackage{hyperref}
\hypersetup{
    colorlinks=true,
    linkcolor=red,
    filecolor=magenta,      
    urlcolor=blue,
    citecolor=purple,
    pdftitle={Forgetting and Imbalance in Robot Lifelong Learning with Off-policy Data},
    pdfpagemode=FullScreen,
    }

\usepackage{caption}
\usepackage{subcaption}
\usepackage{amsfonts}
\usepackage{float}
\usepackage{bbm}
\usepackage{bm}
\usepackage{graphicx}
\usepackage{amsmath}
\usepackage{xcolor} 
\usepackage{wrapfig}

\usepackage{algorithm}
\usepackage{algpseudocode}

\title{Forgetting and Imbalance in Robot Lifelong Learning with Off-policy Data}

\author{Wenxuan Zhou\textsuperscript{1,2}\thanks{Work done as an intern at DeepMind.}, Steven Bohez\textsuperscript{2}, Jan Humplik\textsuperscript{2}, Abbas Abdolmaleki\textsuperscript{2}, Dushyant Rao\textsuperscript{2}, \\\textbf{Markus Wulfmeier\textsuperscript{2}, Tuomas Haarnoja\textsuperscript{2}, Nicolas Heess\textsuperscript{2}} \\
\textsuperscript{1}Robotics Institute, Carnegie Mellon University\\
\textsuperscript{2}DeepMind \\
}
\collasfinalcopy 

\begin{document}

\maketitle
\vspace{-10mm}
\begin{abstract}
Robots will experience non-stationary environment dynamics throughout their lifetime: the robot dynamics can change due to wear and tear, or its surroundings may change over time. Eventually, the robots should perform well in all of the environment variations it has encountered. At the same time, it should still be able to learn fast in a new environment. We identify two challenges in Reinforcement Learning (RL) under such a lifelong learning setting with off-policy data: first, existing off-policy algorithms struggle with the trade-off between being conservative to maintain good performance in the old environment
and learning efficiently in the new environment, despite keeping all the data in the replay buffer. We propose the Offline Distillation Pipeline to break this trade-off by separating the training procedure into an online interaction phase and an offline distillation phase. Second, we find that training with the imbalanced off-policy data from multiple environments across the lifetime creates a significant performance drop. We identify that this performance drop is caused by the combination of the imbalanced \textit{quality} and \textit{size} among the datasets which exacerbate the extrapolation error of the Q-function. 
During the distillation phase, we apply a simple fix to the issue by keeping the policy closer to the behavior policy that generated the data. In the experiments, we demonstrate these two challenges and the proposed solutions with a simulated bipedal robot walking task across various environment changes. We show that the Offline Distillation Pipeline achieves better performance across all the encountered environments without affecting data collection. We also provide a comprehensive empirical study to support our hypothesis on the data imbalance issue.

\end{abstract}

\input{introduction}
\input{related_work}
\input{problem_definition}

\input{pipeline}
\input{imbalance_issue}
\input{experiments}

\input{conclusion}





\bibliography{collas2022_conference}
\bibliographystyle{collas2022_conference}

\newpage
\appendix
\input{appendix}
\end{document}

%% file: math_commands.tex
\usepackage{amsmath,amsfonts,bm}

\def\eqref#1{equation~\ref{#1}}

\def\1{\bm{1}}

\DeclareMathAlphabet{\mathsfit}{\encodingdefault}{\sfdefault}{m}{sl}
\SetMathAlphabet{\mathsfit}{bold}{\encodingdefault}{\sfdefault}{bx}{n}

\DeclareMathOperator*{\argmax}{arg\,max}

%% file: introduction.tex
\section{Introduction}

Lifelong learning, also commonly known as continual learning, studies the problem of learning with a stream of tasks sequentially with incremental, non-stationary data~\citep{thrun1998lifelong, hadsell2020embracing}.
Lifelong learning has been an important topic in artificial intelligence and it naturally reflects the challenges faced by animals and humans~\citep{hassabis2017neuroscience}.
In this work, we study the problem of lifelong robot reinforcement learning 
in the face of changing environment dynamics.
Non-stationary environment dynamics present a practical and important challenge for training reinforcement learning policies on robots in the real world. 
Especially on low-cost or low-tolerance robots, the robot dynamics can change due to wear and tear both during training and deployment.
Also, in most natural settings, the robot's environment will change over time, for instance when the robot encounters new terrains or objects. Ideally, when the robot encounters the same or a similar environment again during deployment, it should be able to draw on the entirety of its past experience to retrieve the previously learned skill. 
In addition, one important property of sequential environment variations in the real world is that the task boundaries may be unknown or not well defined. For example, deformation of the robot can happen gradually. This limits the applicability of many existing approaches in lifelong learning which rely on well-defined task boundaries~\citep{rusu2016progressive, kirkpatrick2017overcoming, lopez2017gradient}. We aim to investigate a practical solution for lifelong robot learning across environment variations without the need of task boundaries.

\begin{figure}
\centering
\vspace{-5mm}
\begin{subfigure}{0.29\textwidth}
    \centering
    \includegraphics[width=0.9\linewidth]{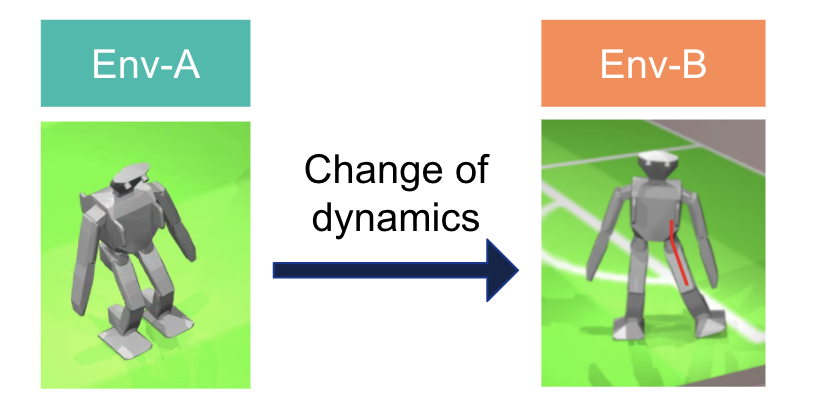}
\end{subfigure}
\begin{subfigure}{0.7\textwidth}
    \centering
    \includegraphics[width=\linewidth]{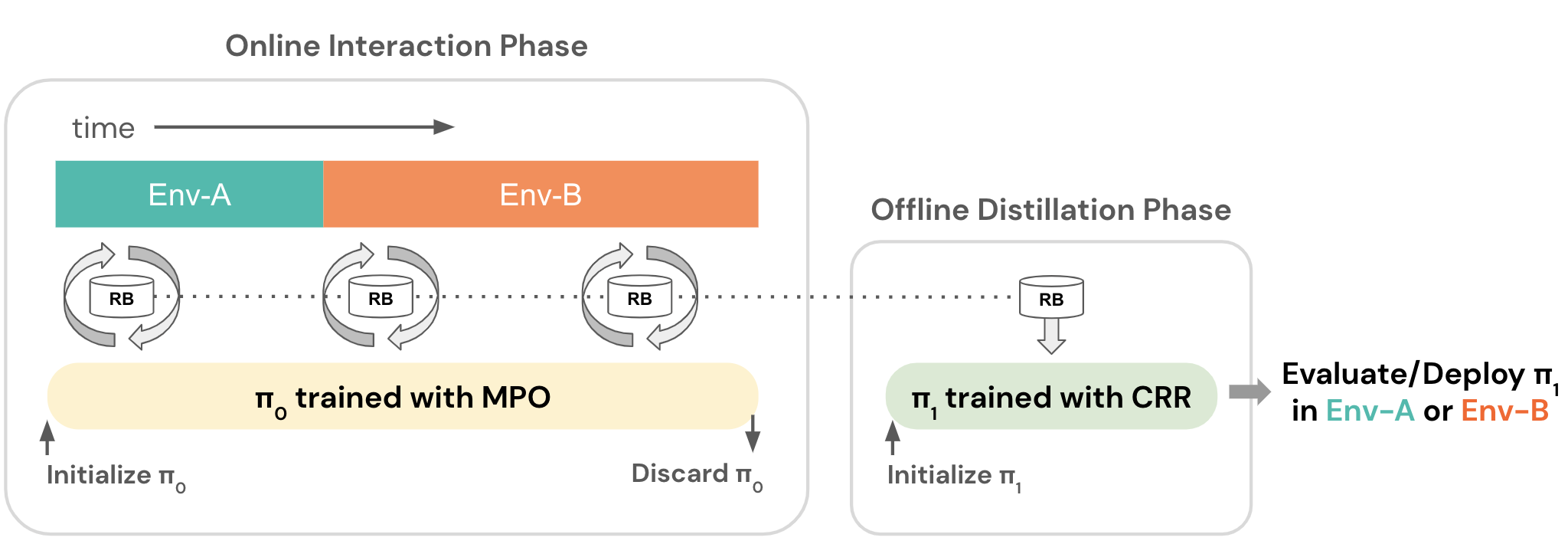}
\end{subfigure}

\caption{We investigate the problem of robot lifelong learning with non-stationary dynamics. For example, we may consider a robot deforms its joint on the right leg at some point during training (left). To obtain a policy that works well on all of the environment dynamics the robot has experienced, we propose the Offline Distillation Pipeline (right): the agent is trained over the environments sequentially during online interaction phase and runs offline distillation at the end of training before deployment. ``RB'' means replay buffer in the figure.}
\vspace{-3mm}
\label{fig:pipeline}
\end{figure}

One important challenge in lifelong learning is the trade-off between remembering the old task (backward transfer) and learning the new task efficiently (forward transfer). The most widely studied aspect of this trade-off is the catastrophic forgetting issue of neural networks~\citep{french1999catastrophic}. We follow the memory-based method to avoid forgetting by simply saving all the incoming data in the replay buffer and train the policy with an off-policy algorithm, which does not require task boundaries~\citep{rolnick2019experience}. However, we find that even if we save all the data across environment variations, ``forgetting'' still happens. In this case, the additional challenge of forgetting is due to the extrapolation error of the Q-function which is widely discussed in the Offline RL literature~\citep{fujimoto2019off}: when the agent does not have access to the previous environments, it becomes ``offline'' over these environments. Thus, the agent cannot correct for the overestimation error of the Q-function by collecting more data in these environments. Conversely, if we use ``conservative'' (or ``pessimistic'') algorithms that force the policy to stay close to the existing replay buffer to maintain the performance in the old environments, it affects data collection and creates difficulties in learning in the new environment~\citep{jeong2020learning}. Different from the stability-plasticity dilemma of neural networks often discussed in the lifelong learning literature, this trade-off between forward and backward transfer is specific to RL due to off-policy data. We propose the \textit{Offline Distillation Pipeline} to disentangle this trade-off into two stages. To learn the task in the latest environment efficiently, we can use any RL algorithm suitable for online data collection without worrying about forgetting. To obtain a policy for deployment that effectively accumulates previous experience across environment variations, we can distill the entire dataset into a policy by treating it as an offline RL problem. 
An illustration can be found in Figure~\ref{fig:pipeline}.

In addition, we investigate a practical consideration of lifelong learning where the stream of experience is imbalanced across environment variations. For example, the agent might be trained on one environment much longer than the other. The ideal lifelong learning algorithm should be robust to such imbalanced experience. In the Offline Distillation Pipeline, we find that training a policy with the imbalanced datasets from multiple environments can sometimes lead to much worse performance than training on each dataset individually. Through the experiments, we provide evidence for the following hypothesis: both the imbalanced quality and the imbalanced size of the datasets become extra sources of extrapolation error in offline learning. The imbalanced quality makes the Q-function biased towards larger values. The imbalanced size leads to more fitting error of the policy network on the smaller dataset, which exacerbates the bootstrapping error caused by out-of-distribution actions. Furthermore, we find that keeping the policy to be closer to the dataset could be a simple yet effective solution to this issue without requiring task boundaries.

In summary, this paper identifies two practical challenges in lifelong robot learning over environment variations and provide corresponding analysis and solutions. The contributions of the paper include the following:
\begin{itemize}
    \item We identify the trade-off between learning in the new environments and remembering the old environments in existing off-policy RL algorithms even when all the data is kept in the replay buffer. We connect this problem to the Offline RL literature and propose the Offline Distillation Pipeline to break this trade-off without the need for task boundaries.
    \item We identify that the dataset imbalance can lead to unexpected performance drop in offline learning and characterize its relationship with extrapolation error with thorough empirical analysis.
\end{itemize}

We evaluate our method on a bipedal robot walking task in simulation with different environment changes. The proposed pipeline is shown to achieve similar or better performance than the baselines across the sequentially changing environments even with imbalanced experience.

%% file: related_work.tex
\section{Related Work}

\textbf{Lifelong Learning:} Lifelong learning has been widely studied in machine learning literature~\citep{thrun1998lifelong, hadsell2020embracing, khetarpal2020towards}. When given a stream of non-stationary data or non-stationary tasks, the agent should maintain the performance of previous tasks (backward transfer) while learning the new task efficiently (forward transfer). One direction of lifelong learning literature focuses more on the issue of backward transfer caused by the catastrophic forgetting of neural networks~\citep{french1999catastrophic}. Existing methods in this direction can be expansion-based~\citep{rusu2016progressive, schwarz2018progress}, regularization-based~\citep{kirkpatrick2017overcoming}, gradient-based~\citep{lopez2017gradient} or memory-based~\citep{rolnick2019experience}. There has also been a line of work in task-agnostic continual learning~\citep{aljundi2019task, aljundi2019gradient, zeno2018task}, where the task boundaries are unknown or not well-defined. We follow the task-agnostic memory-based method from~\citet{rolnick2019experience} by saving all the transitions in the replay buffer. In this paper, we show that there are additional challenges in lifelong reinforcement learning besides the catastrophic forgetting issue of the neural networks. Another direction in lifelong learning focuses on maximizing forward transfer without worrying about forgetting where the performance is only measured by the new task. For example, recent work \citet{xie2021lifelong} studies the problem of learning a sequence of tasks and proposes to selectively use past experience to accelerate forward transfer. 

\textbf{Offline RL:} The additional forgetting issue in off-policy reinforcement learning discussed in this paper is related to offline RL. Thus, the proposed Offline Distillation Pipeline is based on this line of work. Offline RL investigates the problem of learning a policy from a static dataset without additional data collection~\citep{ernst2005tree, lange2012batch, levine2020offline}. Such a problem setting challenges existing off-policy algorithms due to the mismatch between the state-conditioned action distribution induced by the policy and the dataset distribution~\citep{fujimoto2019off}. Previous work has proposed to fix this issue by constraining the policy to be close to the dataset explicitly~\citep{BEAR, BRAC, siegel2020keep, wang2020critic, jeong2020learning} or implicitly~\citep{fujimoto2019off, PLAS_corl2020}, learning a conservative Q-function~\citep{kumar2020conservative}, or modifying the reward based on model uncertainy~\citep{kidambi2020morel, yu2020mopo}. We use Critic Regularized Regression (CRR) from~\citet{wang2020critic} to perform offline distillation.  In terms of related work in imbalanced dataset in offline RL, \citet{zhang2021reducing} investigates imbalanced offline datasets collected by a variety of policies, in contrast to having a mixed dataset from multiple environments in our case. While most of the work in offline RL focuses on one task, \citet{yu2021conservative} studies multi-task offline RL with the goal of improving single task performance by selectively sharing the data across tasks. In contrast, we aim at learning a universal policy for all the tasks which does not rely on task boundaries during training. The difference in the objectives is mainly due to the difference in the domains of interests: the ``tasks'' are defined to have different reward functions in~\citet{yang2022evaluations} while defined to be different dynamics in our case.

\textbf{Distillation:} The proposed pipeline is also related to knowledge distillation~\citep{hinton2015distilling}. In reinforcement learning, policy distillation has been used to compress the network size~\citep{rusu2015policy, schwarz2018progress}, improve multitask learning~\citep{teh2017distral, traore2019discorl}, or improve generalization~\citep{igl2020transient}. In contrast to policy distillation methods which distill the knowledge from networks to networks, we directly distill the data into a policy. This eliminates the need of task boundaries and additional data collection in previous methods. Nonetheless, the proposed pipeline with offline distillation may still share similar benefits of modifying network size or improving generalization because it trains a new policy from scratch~\citep{igl2020transient}.

%% file: problem_definition.tex
\section{Preliminaries}

\subsection{Problem Definition: Lifelong reinforcement learning with environment variations}
\label{prelim_problem}
We define the lifelong learning problem across environment variations to be a time-varying Markov Decision Process (MDP) $\mathcal{M}$ as a tuple $(\mathcal{S},\mathcal{A}, P_t, r,\gamma)$, with state space $\mathcal{S}$, action space $\mathcal{A}$,  non-stationary dynamics function $P_t:\mathcal{S} \times \mathcal{A} \times \mathcal{S} \rightarrow [0,\infty)$ that may change over time $t$, reward function $r: \mathcal{S} \times \mathcal{A} \times \mathcal{S} \rightarrow [r_{min},r_{max}]$, and discount factor $\gamma$. In contrast to our work, most reinforcement learning literature considers static dynamics, which is a special case when $P_t=P$ for all $t$. In reinforcement learning, the objective is to optimize the policy to maximize the return given by $\smash{G_t=\sum_{k=0}^{\infty} \gamma^{k}r(s_{t+k},a_{t+k},s_{t+k+1})}$. We also define a policy $\pi(a|s)$ and its corresponding Q-function $Q^{\pi}(s,a)=\mathbb{E}_{\pi, P_t}[G_t|s_t=s,a_t=a]$ where the expectation is taken over the trajectories that start from an initial state $s$, an initial action $a$, and follow $s_{t+1} \sim P_t(s|s_t, a_t)$ and $a_{t+1} \sim \pi(a|s_{t+1})$ for the following timesteps.

In this work, we assume that the agent experiences ${P_t}$ for a fixed amount of time $[0, T]$ during training, and will be evaluated and deployed at time $T$. This involves efficient data collection across $[0, T]$ and being able to recall the skills at time $T$. We formulate our problem as maximizing the return of the policy over the support of the environment distribution $p(P_{t=[0,T]})$. Intuitively, although the agent might experience one environment more than the other during training, we treat different environments as equally important. For example, in a simplified lifelong learning setting shown in Figure~\ref{fig:pipeline}, the agent experiences $P_t=P_A$ for $t \leq t_A$ and then experiences $P_t=P_B$ for $t_A<t \leq t_A+t_B$. In this case, the objective can be defined as maximizing the performance of $\mathbb{E}_{P_A}[G]+\mathbb{E}_{P_B}[G]$ during evaluation at the end of training.  We aim to learn a policy that works well on both $P_A$ and $P_B$ regardless of $t_A$ and $t_B$. 
Although we have two distinct stages with two environments in this example, in general the task boundaries may not always be accessible or well-defined since ${P_t}$ can change continuously. Without task boundaries, we cannot directly optimize the policy over the support of $P_t$ instead of the density of $P_t$. However, we aim to treat the importance of different environments equally during evaluation.

\subsection{Off-Policy Reinforcement Learning Algorithms}
\label{prelim_algo}
The proposed Offline Distillation Pipeline is built on top of two RL algorithms: Maximum a Posteriori Policy Optimisation (MPO)~\citep{abdolmaleki2018relative, abdolmaleki2018maximum} and Critic Regularized Regression (CRR)~\citep{wang2020critic}. As will be discussed later in Section~\ref{offline_distillation}, our pipeline uses MPO to update the policy during data collection, and uses CRR for offline distillation. 
Although MPO and CRR are both off-policy RL algorithms and share a lot of similarities, they are designed for different problem settings. MPO works well in the online setting, i.e.\ when data collection is allowed. CRR is designed for offline reinforcement learning, i.e.\ to learn from a fixed dataset without additional data collection.  To stabilize learning in the offline setting, CRR attempts to avoid selecting actions outside of the dataset, which also renders it more ``conservative''. More discussion of the connections between CRR and MPO can be found in~\citet{jeong2020learning,abdolmaleki2021multi}. Both algorithms alternate between policy evaluation and policy improvement. Both algorithms perform \textbf{policy evaluation} to estimate the Q-function $\hat{Q}^{\pi}(s_t,a_t)$ using the Bellman Operator $\mathcal{T}$:

\begin{equation}
\label{bellman}
\mathcal{T} \hat{Q}^{\pi}(s_t,a_t)=\mathbb{E}_{s_{t+1}, a_{t+1} \sim \pi(a|s_{t+1})}[r_t+\gamma \hat{Q}^{\pi}(s_{t+1}, a_{t+1})],
\end{equation} 

but they differ in the \textbf{policy improvement} step.



\textbf{Maximum a Posteriori Policy Optimisation (MPO):}
To improve the current policy $\pi_{old}(a|s)$ given the corresponding Q-function $Q^{old}(a,s)$ and a state distribution $\mu(s)$, MPO performs two steps. In the first step, for each state $s \sim \mu(s)$, an improved policy $q(a|s) \propto \pi_{old}(a|s) f({Q}^{old}(s, a))$ is obtained where $f$ is a transformation function that gives higher probabilities to actions with higher Q-values. In the second step, a new parametric policy $\pi_{new}$ is obtained by distilling the improved policies $q(a|s)$ into a new parametric policy using the supervised learning loss: 

\begin{equation}
\label{mpo_policy_improvement}
\pi_{new} = \argmax_\pi \int \mu(s) \int q(a|s) \log \pi(a|s) f({Q}^{old}(s, a)) da ds
\end{equation}

In practice, we represent $q(a|s)$ as a non-parametric policy consists of samples from $\pi_{old}(a|s)$ for each state and re-weighting each sample by $f({Q}^{old}(s, a))$. If the exponential function is chosen as the transformation function $f$, the improved policy can be written as $q(a|s) \propto \pi_{old}(a|s) \exp({Q}^{old}(s, a)/\beta)$ where $\beta$ is the temperature term. This is the solution to the following KL regularized RL objective that keeps the improved policies $q$ close to the current policy $\pi_{old}$ while maximising the expected Q-values: 

\begin{equation}
\label{mpo_objective}
q = \argmax_q \mathbb{E}_{s \sim \mu(s)}[ \mathbb{E}_{a \sim q(\cdot|s)} [Q^{old}(s,a)] - \beta \mathrm{KL}(q(\cdot|s)||\pi_{old}(\cdot|s))]
\end{equation}

\textbf{Critic Regularized Regression (CRR):} CRR follows a similar procedure as MPO for the policy improvement step. The major difference is the way of constructing the improved policy $q$. Since CRR is designed for offline RL, the optimization objective is to improve the policy according to the Q-function while keeping the policy close to the dataset distribution.  In CRR, we construct the improved policies based on the joint distribution of $\mu_{\mathcal{B}}(a,s)$ by sampling state-action pairs from the dataset $\mathcal{B}$. Thus, the improved policy for each state $s \sim \mu_\mathcal{B}(s)$ is defined as a joint distribution $q(a,s) \propto \mu_{\mathcal{B}}(a,s) f({Q}^{old}(s, a))$ instead of a conditional distribution $q(a|s)$ as in MPO. Similarly, Equation~\ref{mpo_policy_improvement} can be modified to obtain a new parametric policy $\pi_{new}$:

\begin{equation}
\label{crr_policy_improvement}
\pi_{new} = \argmax_\pi  \int q(a,s) \log \pi(a|s) das
\end{equation}

When the transformation function is the exponential function, the improved policy can be written as $q(a,s) \propto \mu_{\mathcal{B}}(a,s) \exp({Q}^{old}(s, a)/\beta)$ which is a solution to the following objective similar to Equation~\ref{mpo_objective}:

\begin{equation}
\label{crr_objective}
q = \argmax_q \mathbb{E}_{s,a \sim q}[\hat{Q}^\pi(s,a)] - \mathbb{E}_{s,a \sim \mathcal{B}}[\beta \mathrm{KL}[q(a,s)||\mu_\mathcal{B}(a,s)]].
\end{equation}

In Equation~\ref{crr_objective}, when the temperature $\beta$ is higher, the constraint on staying close to the dataset becomes stronger, which makes the policy more ``conservative''. A common practice is to replace the Q-value by the advantage in the transformation function: $q(a,s) \propto \mu_\mathcal{B}(a, s) \exp({A}^{old}(s,a)/\beta)$. 
Besides the exponential function, another popular choice of the transformation is the indicator function: $q(a,s) \propto \mu_\mathcal{B}(a, s) \mathbbm{1}[{A}^{old}(s, a)>0]$ where $A$ is the advantage function. The indicator function corresponds to an exponential transformation clipped to $[0,1]$ with $\beta \to 0$ which is less ``conservative''.

%% file: pipeline.tex
\newpage
\section{Forward and Backward trade-off in Lifelong Reinforcement Learning}
\label{tradeoff}

\begin{wrapfigure}{r}{0.25\textwidth}
    \centering
    \vspace{-5mm}
    \includegraphics[width=\linewidth]{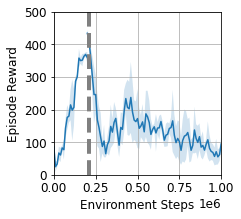}
   \vspace{-3mm} \caption{Forgetting in MPO: The figure shows a performance drop in the original environment after a switch of environment at 200k steps.}
    \label{fig:forgetting}
    \vspace{-2mm}
\end{wrapfigure}

To build a pipeline for lifelong learning, we need to first deal with the catastrophic forgetting issue of neural networks, as widely discussed in the literature~\citep{hadsell2020embracing}. We follow the memory-based approach from \citet{rolnick2019experience} by saving all the transitions across the agent's life-cycle and run off-policy algorithms such as MPO~\citep{abdolmaleki2018maximum}. In off-policy algorithms, the policy is used for exploration in the latest environment while being trained on the entire history of data. However, we still observe that ``forgetting'' happens in the old environment following this setup.  Figure~\ref{fig:forgetting} shows an example in which the policy experiences a change in the environment dynamics at 200k steps while being evaluated in the first environment across the full training process. More details of the experiment can be found in Section~\ref{setup}. Once the policy starts training in a new environment, the performance of MPO drops significantly even if the data from the old environment is kept in the replay buffer. We identify this extra challenge in lifelong reinforcement learning besides the catastrophic forgetting issue of the neural network.

The reason behind this drop is related to the issues of applying off-policy algorithms to Offline Reinforcement Learning problems~\citep{fujimoto2019off}. The objective of offline RL is to learn a policy from a fixed dataset without further exploration. During training, due to the extrapolation error in the Q-function, the policy might select overestimated actions beyond the dataset. This error will be accumulated by bootstrapping during Q-function updates which results in significant overestimation bias of the Q-function. When the agent does not have access to collect more data to correct the overestimation bias, the performance of the policy will drastically degrade. Thus, off-policy algorithms designed with the assumption of active data collection often break under this problem setting. Similarly, in the lifelong learning scenario discussed above, when the agent switches from one environment to another, it is essentially training over the static dataset of the old environment. When the agent cannot collect more data in the old environment, it cannot correct the extrapolation error on those state-action pairs.

\begin{wrapfigure}{r}{0.25\textwidth}
    \vspace{-6mm}
    \centering
    \includegraphics[width=\linewidth]{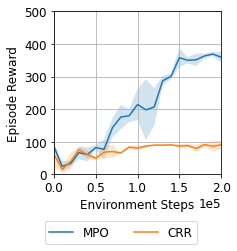}
   \vspace{-4mm} \caption{CRR is not as efficient as MPO when training from scratch.}
    \label{fig:crr_online}
    \vspace{-5mm}
\end{wrapfigure}

Prior work in offline RL proposes to fix the overestimation issue of off-policy algorithms by restricting the policy $\pi(a|s)$ to be closer to the conditional distribution $\mu_{\mathcal{B}}(a|s)$ of the dataset, such as Critic Regularized Regression (CRR) as described in Section~\ref{prelim_algo}. If we apply a similar ``conservative'' objective in the lifelong learning setting, we find that it is able to reduce the forgetting issue. However, it will instead affect forward transfer due to the conservatism. Figure~\ref{fig:crr_online} shows an example of running CRR from scratch, which can be viewed as the beginning stage of a lifelong learning experiment. Although CRR has been shown to have strong performance on offline RL benchmarks, it does not have good performance when exploration is needed due to the constraint on the policy. In the experiment, we will further show that tuning the constraint will lead to either forgetting or ineffective forward transfer.

The above examples demonstrate the trade-off between preserving performance in the old environment (backward transfer) and exploring in the new environment effectively (forward transfer). Note that this is different from the ``stability-plasticity'' dilemma in previous lifelong learning literature in two ways. First, in terms of backward transfer, the issue of forgetting rises from the extrapolation error of Q-function which is specific to off-policy reinforcement learning. Second, in terms of forward transfer, previous work mainly considers the trade-off between past and \textit{recent} experience from the streaming data. The issue we discuss above is a trade-off between past and \textit{future} experience which is specific to reinforcement learning where the performance highly depends on effective data collection. 

\section{Offline Distillation Pipeline}
\label{offline_distillation}

To address this trade-off, we propose the Offline Distillation Pipeline shown in Figure~\ref{fig:pipeline}. During data collection across environment variations, we can use any RL algorithm that maximizes forward transfer without considering forgetting. At the end of training, we ``distill'' the experience into a single policy by treating the entire dataset as an offline RL dataset. In this paper, we use MPO to train the policy for data collection, and use CRR during offline distillation. In this way, the forgetting issue of the off-policy data is handled by the distillation step without affecting exploration. The full algorithm can be found in Appendix~\ref{appd:algo}.

There are several benefits of this pipeline that are especially important for lifelong learning of real robots. First, the proposed pipeline does not require task boundaries. The wear and tear of the robot might happen over time and sometimes the change of the environment might not be immediately noticeable. This is different from a common multi-task learning setting where the task switches are well defined (such as learning to stand up and then learning to walk). In our method, the distillation step happens after the agent has experienced the sequence of environments and treats the replay buffer as a single dataset. Second, our method is flexible on the choice of data collection methods since the offline distillation phase only shares the replay buffer dataset with the online interaction phase. For example, the training of multiple robots can happen in parallel or sequentially, and potentially with different choices of algorithms. The Offline Distillation Pipeline can reuse all of these previous experience within the ``lifetime'' of the platform. 



%% file: imbalance_issue.tex
\section{Imbalanced Experience in Offline Distillation}
\label{imbalance_issue}

The second challenge we identify in such a lifelong learning setting is the issue of imbalanced experience with offline data. In the offline distillation phase, we find that when the policy is trained over the combined dataset from multiple environments, the imbalance of the datasets might create an unexpected performance drop. For example, following Figure~\ref{fig:pipeline}, the agent is first trained in Env-A and then switches to Env-B. During the offline distillation phase, we use CRR to train a policy with the combined dataset $\mathcal{D}_A \cup \mathcal{D}_B$ and evaluate the performance in both environments, as formulated in Section~\ref{prelim_problem}. We find that this sometimes results in worse performance in Env-A compared to training on $\mathcal{D}_A$ alone. Although previous work has studied the problem of data imbalance in supervised learning~\citep{johnson2019survey, ren2018learning}, the issue we observe has the extra complexity from the boostrapping procedure in off-policy RL. We provide evidence to the following hypothesis: Both the imbalanced \textbf{quality} and the imbalanced \textbf{size} of the combined dataset lead to additional extrapolation error of the Q-function in offline learning which contribute to the performance drop. As we discussed in Section~\ref{tradeoff}, extrapolation error of the Q-function plays an important role in the failure cases in offline RL. The imbalanced dataset exacerbates this problem in the following way: if one dataset has a higher average return than the other, it may cause overestimation bias of the Q-function for the ``weaker'' dataset. At the same time, if there is a large size imbalance, the policy network will be trained with more data points from one environment than the other. In this way, the policy may create more out-of-distribution actions in the environment that comes with a smaller dataset which makes the extrapolation error worse. Both of these two aspects contribute to the undesirable performance we observe in offline distillation phase. In the experiment, we provide evidence to support this hypothesis and eliminate other potential factors. 

To build a robust algorithm for lifelong learning, we need to improve the offline distillation phase to achieve good performance on all of the environments despite imbalanced experience. We prefer a solution that does not rely on task boundaries as discussed before. Our insight is that since both the quality imbalance and the size imbalance eventually result in additional extrapolation error, we can follow the conservative objective in offline RL and make the policy even more conservative to compensate for this issue. As shown in Equation~\ref{crr_objective}, the temperature $\beta$ controls the strength of the KL term in the policy improvement objective in CRR. With a larger temperature, the policy is constrained to be closer to the behavior policy $\mu_{\mathcal{B}}(a|s)$ of the dataset. We find that the imbalanced dataset requires a higher strength of the KL term compared to single dataset training to compensate for the additional extrapolation error. In the experiment, we show that increasing $\beta$ is a simple yet effective fix to the data imbalance problem. The effectiveness of increasing $\beta$ can also serve as an evidence that the performance drop is highly related to extrapolation error. Note that increasing $\beta$ only makes the policy more ``conservative'' during the distillation phase which will not affect exploration.

%% file: experiments.tex
\section{Experiments}
\label{experiments}

\subsection{Experiment Setup}
\label{setup}
We study the lifelong learning problem in a simulated bipedal walking task, where the goal is to maximize the forward velocity while avoiding falling. Our experiments involve a small humanoid robot, called OP3\footnote{\url{https://emanual.robotis.com/docs/en/platform/op3/introduction/}}, that has 20 actuated joints and has been previously used to train walking directly on hardware~\citep{bloesch2022towards}. All of the experiments in this work are conducted in simulation both due to limited access to hardware and for a more controlled experiment setting. 
All of the results are averaged over 3 random seeds. The shaded area of the curves and the error bars of the bar plots represent $\pm$ one standard deviation across seeds.

The experiments in the section are based on the setup where the robot is trained in Env-A for 0.2M steps, and then trained in Env-B for 1M steps (Figure~\ref{fig:pipeline}). The goal is to achieve good performance at the end of training in both Env-A and Env-B. Additional experiment setups with three environments and parallel sharing can be found in Appendix~\ref{appd:3envs}. To evaluate the generality of the results, we consider different types of changes in the environment including softer ground texture, hip joint deformation and larger foot size (Figure~\ref{fig:setup}). The parameters for each change of the environment are chosen to create a clear performance drop when we perform zero-shot transfer of a policy trained in the default environment to the new environment. In the following experiments when there is a switch from Env-A to Env-B, we use the default environment as Env-A, and change one of the physical parameters to create Env-B. When we switch from one environment to another, we always keep the previous policy, Q-function and the replay buffer. 

To remove partial observability in non-stationary dynamics, we include the ground truth physical parameters in the observation. This eliminates the possibility that the issue we observe in the lifelong learning pipeline and the imbalanced experience are caused by the partial observability. The results we provide can be served as an upper bound on the expected performance without the physical parameters. However, the physical parameters are not explicitly used to denote task boundaries in the proposed method. In a realistic scenario, the partial observability could be handled by memory or system identification methods~\citep{yu2017preparing, heess2016learning, zhou2019environment}. In these cases, the variations of the environments might be represented as continuous embeddings which cannot be used as task boundaries. Thus, we avoid relying on the physical parameters in the proposed pipeline as task boundaries.

\subsection{Offline Distillation for Lifelong Reinforcement Learning}
\label{result:pipeline}

\begin{figure}[]
    \vspace{-5mm}
   \begin{minipage}{0.42\textwidth}
     \centering
     \includegraphics[width=\linewidth]{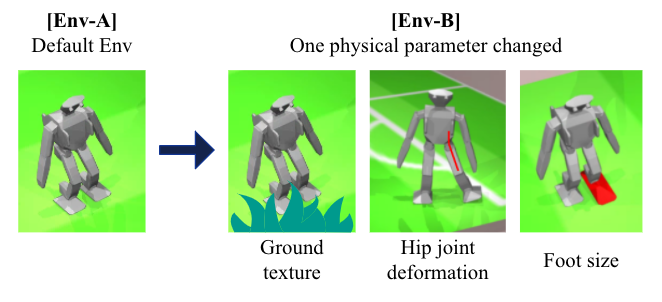}
     \vspace{-6mm}
     \caption{Experiment Setup: A bipedal robot walking task with various environment variations.}\label{fig:setup}
   \end{minipage}\hfill
   \begin{minipage}{0.55\textwidth}
     \centering
    \includegraphics[width=\linewidth]{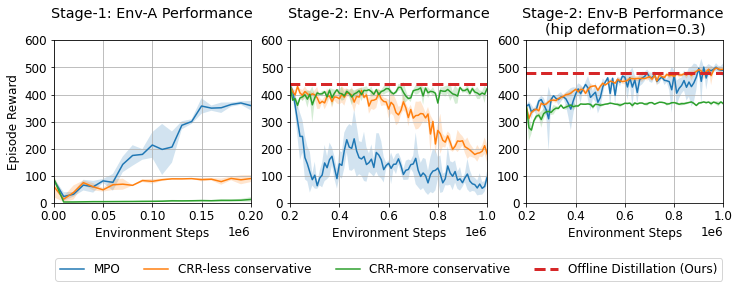}
    \vspace{-6mm}
     \caption{Evaluation curves of the lifelong learning experiment across two stages. The Offline Distillation Pipeline effectively breaks the trade-off between forward and backward transfer and achieves better performance than the baselines at the end.}\label{fig:online}
   \end{minipage}
\end{figure}

In this section, we demonstrate the trade-off between forward transfer and backward transfer in lifelong reinforcement learning and the effectiveness of the Offline Distillation Pipeline. The policy is trained from scratch in Env-A during Stage-1 and then switched to Env-B during Stage-2 (as illustrated in Figure~\ref{fig:pipeline}). We compare the performance of MPO, CRR with a less conservative objective (with an indicator function which corresponds to $\beta \to 0$ as discussed in \ref{prelim_algo}), CRR with a more conservative objective ($\beta=1$) and the Offline Distillation Pipeline. In the example shown in Figure~\ref{fig:online}, the robot in Env-B has its right hip joints deformed for 0.3 rad. The results for more Env-B variations are included in Appendix~\ref{appd:pipeline}. 

We first demonstrate the forward transfer problem in Stage-1. As shown on the left figure in Figure~\ref{fig:online}, conservative algorithms such as CRR cannot learn as efficiently as MPO from scratch. Then, we show the backward transfer problem in Stage-2. To compare the performance drop better, we enforce the same starting performance of Stage-2 for all the baselines. This is done by loading the same agent (networks and replay buffer) trained with MPO from Stage-1. All the baselines keep training on the loaded replay buffer with Env-A data while collecting new data in Env-B with the latest policy. In Stage-2, the policy only collects data in Env-B, but we evaluate the performance in both Env-A and Env-B. From the middle figure in Figure~\ref{fig:online}, the performance of MPO drops significantly in Env-A after the switch. This could potentially be explained by the fact that Env-A transitions are ``offline'' during Stage-2 and thus the extrapolation error starts to accumulate. The forgetting issue also happens in the less conservative CRR, despite being less severe than MPO. The more conservative CRR keeps the performance of Env-A effectively which indicates that the performance drop is indeed related to the extrapolation issue in offline RL. However, as shown in the right figure in Figure~\ref{fig:online}, more conservative CRR does not improve the performance in Env-B as the other baselines. 

In summary, these baselines either struggle with backward transfer or forward transfer. As described in Section~\ref{offline_distillation}, we propose the Offline Distillation Pipeline which distills the data collected by MPO using CRR. In this experiment, we perform the distillation step at the end of Stage-2. The performance is shown as the dotted lines in the figures. Taking the best of both world, our method can achieve better performance than the baselines in both environments. Note that in this experiment we include the results of the proposed method with the data imbalance issue fixed which will be discussed in the next section. 

\subsection{Imbalanced Experience in Offline Distillation}

\label{result:imbalance}
As discussed in Section~\ref{imbalance_issue}, we sometimes observe a performance drop during the distillation phase in the proposed pipeline with imbalanced experience. We will provide experimental evidence for the hypothesis that the decrease in performance is caused by the imbalanced size and quality between the datasets. In this section, we use CRR with the indicator function by default which corresponds to a less conservative CRR objective as discussed in Section~\ref{prelim_algo}.

\begin{figure}
\centering
\begin{subfigure}{0.49\textwidth}
    \includegraphics[width=\linewidth]{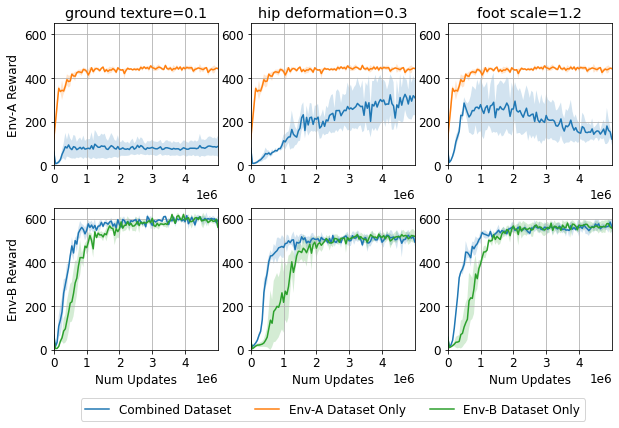}
    \caption{Reward}
    \label{fig:reward}
\end{subfigure}
\begin{subfigure}{0.49\textwidth}
    \includegraphics[width=\linewidth]{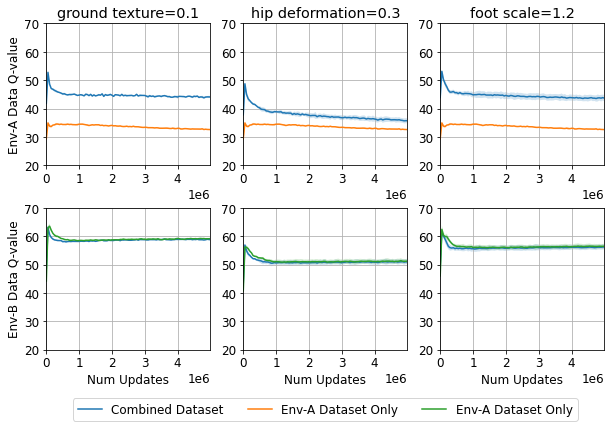}
    \caption{Q-values}
    \label{fig:qvalue}
\end{subfigure}
\vspace{-1mm}
\caption{The imbalance issue in Offline Distillation: Training CRR with the combined dataset results in much lower performance in Env-A compared to training on the individual dataset. However, the worse performance corresponds to a higher average Q-value which indicates overestimation.}
\vspace{-5mm}
\label{fig:imbalance}
\end{figure}

\begin{figure}
    \centering
    \vspace{-4mm}
    \includegraphics[width=0.9\linewidth]{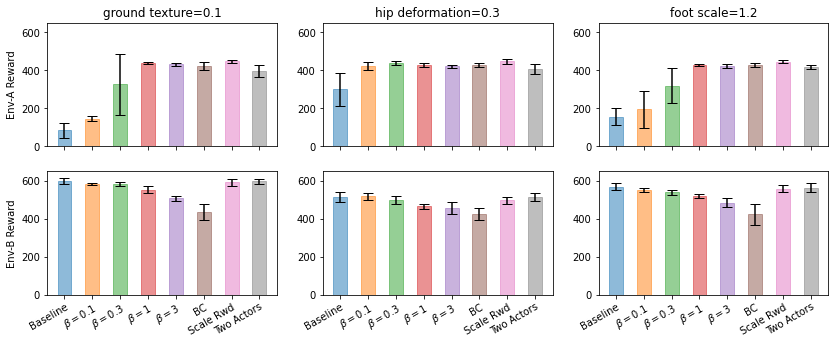}
    \vspace{-2mm}
    \caption{To study the imbalance issue, the figure shows the final performance of different variants of the offline distillation step at 5e6 policy updates. The \textbf{Baseline} corresponds to the performance of \textbf{Combined Dataset} in Figure~\ref{fig:imbalance}.}   
    \label{fig:barplots}
\end{figure}

\textbf{The Imbalance Issue.} 
Following the setup in the previous section (Figure~\ref{fig:pipeline}), we run the offline distillation step at the end of Stage-2 with the combination of dataset $\mathcal{D}_A$ in Env-A from Stage-1 and $\mathcal{D}_{B_i}$ in Env-B from Stage-2, both collected by MPO. We apply different environment variations in Env-B to generate $\mathcal{D}_{B_i}$ with different $i$ while keeping $\mathcal{D}_A$ the same in the combined dataset (Figure~\ref{fig:setup}). In Figure~\ref{fig:reward}, the blue curves (\textbf{Combined Dataset}) show the performance of the distilled policy $\pi_{\mathcal{D}_A \cup \mathcal{D}_{B_i}}$ evaluated in Env-A and Env-B across the CRR training process of the distillation stage. Each column corresponds to a different $\mathcal{D}_{B_i}$. Note that there is no data collection in this stage and the x-axis here is the number of policy updates in CRR rather than environment steps. As a comparison, we run CRR on $\mathcal{D}_A$ and $\mathcal{D}_{B_i}$ separately with the same hyperparameters (\textbf{Env-A Dataset Only}, \textbf{Env-B Dataset Only}). Given that there is no partial observability (see Section~\ref{setup}), we expect $\pi_{\mathcal{D}_A \cup \mathcal{D}_{B_i}}$ to have similar or better performance than $\pi_{\mathcal{D}_A}$ evaluated in Env-A, and $\pi_{\mathcal{D}_{B_i}}$ evaluated in Env-$B_i$. From the second row of Figure~\ref{fig:reward}, the performance in Env-$B_i$ is similar between $\pi_{\mathcal{D}_A \cup \mathcal{D}_{B_i}}$ and $\pi_{\mathcal{D}_{B_i}}$ at convergence. 
However, from the first row of Figure~\ref{fig:reward}, the performance of $\pi_{\mathcal{D}_A \cup \mathcal{D}_{B_i}}$ in Env-A is much worse than training with $\mathcal{D}_{A}$ alone: we observe the blue curves to converge at a lower reward, converge much slower, or becomes unstable during training. Despite being trained on the same $\mathcal{D}_A$, the distilled policies $\pi_{\mathcal{D}_A \cup \mathcal{D}_{B_i}}$ have very different performance in Env-A due to the fact that they are combined with different $\mathcal{D}_{B_i}$. Although the performance drop in Env-A does not always happen, it is important to understand when and why the performance degrades to develop a robust lifelong learning pipeline that works for diverse settings. To get more insights of the problem, we also train a behavior cloning policy with the combined dataset. Figure~\ref{fig:barplots} includes the final performance of CRR  (\textbf{Baseline}) and behavior cloning (\textbf{BC}) over the combined dataset. Despite the size imbalance, with only supervised learning, BC performs reasonably well in Env-A. The CRR Baseline is much worse than BC in Env-A. This comparison indicates that the performance drop we observe in Env-A is more likely to be rooted in the RL procedure, instead of being a regular data imbalance problem in a supervised setting. We have also tried a few sanity check experiments including increasing batch size, increasing network capacity, or using a mixture of Gaussians as the policy output. None of these can prevent the performance drop in Env-A. In the following sections, we will discuss the most important experiments that can support our hypothesis discussed in Section~\ref{imbalance_issue}. 

\textbf{Overestimation due to the imbalanced quality.} Figure~\ref{fig:qvalue} plots the average estimated Q-values of each method over $(s, a) \sim \mathcal{D}_A$ and $(s, a) \sim \mathcal{D}_{B_i}$ separately. Although $\pi_{\mathcal{D}_A \cup \mathcal{D}_B}$ has a lower performance in Env-A than $\pi_{\mathcal{D}_A}$, the corresponding Q-functions $Q^{\pi_{\mathcal{D}_A \cup \mathcal{D}_{B_i}}}$ produce higher estimated Q-values over Env-A datapoints, which indicates significant overestimation. In contrast, if we compare the Q-values over $(s, a) \sim \mathcal{D}_{B_i}$ on the second row, the curves are similar to each other within each plot. This indicates that $Q^{\pi_{\mathcal{D}_A \cup \mathcal{D}_{B_i}}}$ suffers from overestimation specifically for Env-A data points. Furthermore, we observe that the average Q-value over $\mathcal{D}_{B_i}$ is higher than $\mathcal{D}_{A}$ for the individual dataset experiments. This is because $\mathcal{D}_{A}$ is collected by training from scratch, while $\mathcal{D}_{B_i}$ is collected during Stage-2 where the policy is bootstrapped from the previous experience and starts from a higher performance (see Figure~\ref{fig:online}). This observation leads us to the hypothesis that the high value datapoints in Env-B bias the Q-function which leads to overestimation for Env-A datapoints. To verify this hypothesis, we perform an experiment where we scale the reward for all the transitions in $\mathcal{D}_B$ by $0.5$, which does not change the optimal solution of the policy. After this change, the distilled policy with the combined dataset works well on both environments (\textbf{Scale Rwd} in Figure~\ref{fig:barplots}) and achieves similar performance as the individual dataset baselines. 
However, re-scaling the reward is not an acceptable solution in our problem setting because it requires the knowledge of task boundaries. It only serves as an analysis to demonstrate the imbalance issue due to the quality of the dataset.

\textbf{Additional fitting error of the actor.} We also test the contribution of the fitting error of the actor to the overall extrapolation error.
We use separate actor networks for each dataset when training CRR on the combined dataset $\mathcal{D}_A \cup \mathcal{D}_{B_i}$ (\textbf{Two Actors} in Figure~\ref{fig:barplots}): Actor-A and Actor-B are trained with the transitions from Env-A and Env-B respectively, while the critic is shared across two environments. This change also makes the policy work well in both environments despite that the imbalanced reward is not corrected. In the Two Actors experiment, we find that the overestimation over $\mathcal{D}_{A}$ still exists but has been reduced. Together with the Scale Reward experiment, the results indicate that the overestimation we observe in Figure~\ref{fig:qvalue} in Env-A is caused by two sources of error: the imbalanced quality creates overestimation; The imbalanced size creates more fitting error of the actor which results in more out-of-distribution actions that may take advantage of the overestimation. Note that using two separated actors also requires task boundaries and only serves as an analysis.

\begin{figure}
    \centering
    \includegraphics[width=0.75\linewidth]{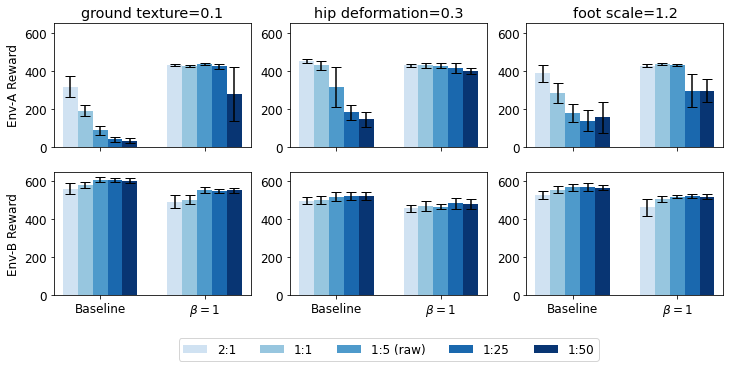}
    \vspace{-2mm}
    \caption{Sensitivity analysis of data imbalance: Higher $\beta$ makes CRR more robust to different ratios of dataset imbalance. The legend indicates the dataset size ratio of Env-A:Env-B corresponding to different colors.}
    \label{fig:sampling_ratio}
    \vspace{-4mm}
\end{figure}

\textbf{Effectiveness of the temperature.} As shown in previous sections, fixing either the imbalanced quality or the fitting error of the actor makes the algorithm stable when evaluated in both Env-A and Env-B. However, we need a solution that does not require task boundaries. As proposed in Section~\ref{imbalance_issue}, increasing the temperature term $\beta$ in CRR can largely fix this issue. Figure~\ref{fig:barplots} includes the performance of CRR with different $\beta$. \textbf{Baseline} uses an Indicator function as the transformation function which corresponds to very small $\beta$. With increased $\beta$, the performance in Env-A increases. Although we observe a minor drop in Env-B with high $\beta$, the overall performance in both Env-A and Env-B are reasonably satisfactory. To further demonstrate the effectiveness of increasing $\beta$, we conduct an experiment where we upsample either $\mathcal{D}_A$ or $\mathcal{D}_{B_i}$ to simulate other compositions of the combined dataset (Figure~\ref{fig:sampling_ratio}). As mentioned in Section~\ref{setup}, the size ratio of $\mathcal{D}_A:\mathcal{D}_{B_i}$ is $1:5$ (denoted as \textbf{raw}). The performance in Env-A of CRR with the indicator function (\textbf{Baseline}) decreases drastically with higher Env-B sampling ratio. Interestingly, when the size ratio is $1:1$, the policy is still not able to consistently achieve the single dataset performance in Env-A (which is expected to be above 400 as shown in Figure~\ref{imbalance_issue}). In contrast, CRR with $\beta=1$ works well across a wider range of size ratios (which is what we use in Section~\ref{result:pipeline}). As shown in previous work~\citep{wang2020critic}, the specific choice of $\beta$ could be domain-dependent. The more important takeaway from this analysis is that if we observe a performance drop during RL training with an imbalanced dataset, we may consider increasing the conservativeness of the policy to compensate for the additional extrapolation error, such as increasing $\beta$ in CRR.

%% file: conclusion.tex
\section{Conclusion}

In this work, we investigate the lifelong learning problem of variations in environment dynamics as commonly observed when learning on robot hardware. Our main contributions include identifying and addressing two challenges within such a problem setting: First, we find that there is a trade-off between backward and forward transfer of existing RL algorithms in this problem setting even when we keep all of the transitions in the replay buffer. We connect the problem to offline RL and propose the Offline Distillation Pipeline to break this trade-off. In the proposed pipeline, the forgetting issue is prevented by distilling the replay buffer data across multiple environments into a universal policy as an offline RL problem. In this way, the solution to the forgetting problem is disentangled from data collection. We empirically verify the effectiveness of the pipeline through a bipedal robot walking task in simulation across various physical changes. Second, we identify an potential issue with imbalanced experience in offline distillation. Through controlled experiments, we demonstrate how the quality imbalance and the increased fitting error of the actor might exacerbate extrapolation error and create a performance drop. We also provide a simple yet effective solution to this issue by increasing the temperature term in CRR.

The insights from this work could potentially be applied in other settings beyond the lifelong learning problem of varying dynamics. For example, the Offline Distillation Pipeline can be used in other lifelong reinforcement learning settings with a different definition of ``task'' without varying dynamics. The imbalance issue may also happen in other cases of multi-task learning in offline RL, or in single-task RL with sufficient non-stationarity (e.g. due to partial observability). In future work, we hope to see the proposed method being verified and deployed in more settings including having multiple distillation steps across the training procedure or real robot experiments.

%% file: appendix.tex
\section{Additional results on the offline distillation pipeline}
\label{appd:pipeline}

\begin{figure}[H]
\centering
\includegraphics[width=0.6\linewidth]{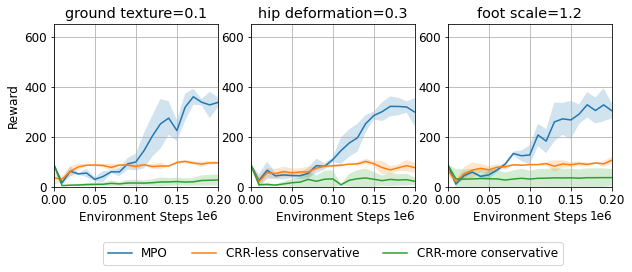}
\caption{Comparison of off-policy algorithms for training from scratch which corresponds to the beginning stage of a lifelong learning experiment. This is an extension of Figure~\ref{fig:online} Stage-1 result.}\label{fig:appd_stage1}
\end{figure}

\begin{figure}[H]
\centering
\includegraphics[width=0.7\linewidth]{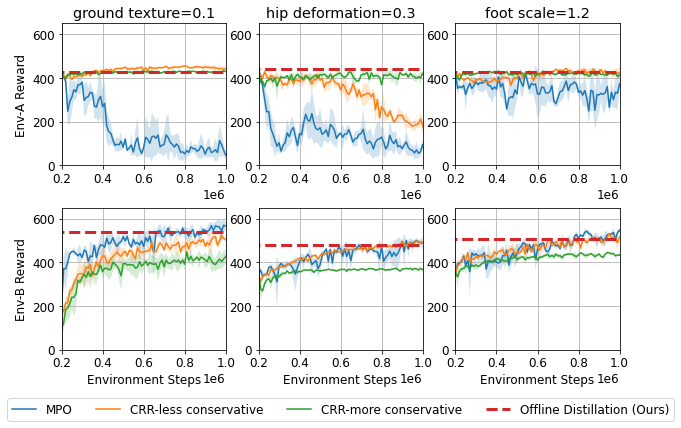}
\caption{Comparison of off-policy algorithms during Stage-2 of the lifelong learning experiment.  This is an extension of Figure~\ref{fig:online} Stage-2 results.}\label{fig:appd_stage2}
\end{figure}

In Figure~\ref{fig:online}, we present the results of the two-stage lifelong learning experiments when Env-A is the default environment and Env-B has the hip joint of the robot deformed by 0.3 rad. In this section, we include more results across more environment variations. To demonstrate the difficulty of data collection with conservative algorithms, Figure~\ref{fig:appd_stage1} shows the performance of each algorithm when they are trained from scratch, which corresponds to the first stage of the lifelong learning setup discussed in Section~\ref{result:pipeline}. MPO performs significantly better than both versions of CRR. Figure~\ref{fig:appd_stage2} shows the performance during Stage-2 where all of the algorithms loaded an agent which is pretrained in Env-A (the default environment) for 0.2M steps. The Offline Distillation Pipeline can achieve the best performance consistently across different Env-B variations. 

\newpage
\section{Algorithm}
A summary of the algorithm is included in Algorithm~\ref{algo}.

\label{appd:algo}

\begin{algorithm}
\caption{Offline Distillation Pipeline}\label{algo}
\begin{algorithmic}
\Require Maximum data collection step $N$. Maximum offline traning step $M$.  A training environment with non-stationary dynamics $P_t$. A testing environment with dynamics $P_{test}$ within the support of the training distribution. \\

\State \textbf{// Online Interaction Phase}
\State Initialize policy $\pi_0$ and Q-function $Q_0$. Initialize an empty replay buffer $D$.
\For{\texttt{$t = 1...N$}}
\State Collect data with the current dynamics $P_t$ and save the transition $(s_t, a_t, s_{t+1})$ into the replay buffer $D$.
\State Sample a batch of transitions $\{(s_i, a_i, s_{i+1})\}_i$ from $D$.
\State Update $Q_0$ with MPO Policy Evaluation Step according to  Equation~\ref{bellman}.
\State Update $\pi_0$ with MPO Policy Improvement Step according to Equation~\ref{mpo_policy_improvement}.
\EndFor
\\
\State \textbf{// Offline Distillation Phase}
\State Initialize policy $\pi_1$ and Q-function $Q_1$.
\For{\texttt{$t = 1...N$}}
\State Sample a batch of transitions $\{(s_i, a_i, s_{i+1})\}_i$ from $D$.
\State Update $Q_1$ with CRR Policy Evaluation Step according to  Equation~\ref{bellman}.
\State Update $\pi_1$ with CRR Policy Improvement Step according to Equation~\ref{crr_policy_improvement}.
\EndFor
\\
\State \textbf{// Deployment}
\State Execute the distilled policy $\pi_1$ in the testing environment $P_{test}$.

\end{algorithmic}
\label{algo}
\end{algorithm}

\section{Additional results with three environments and parallel sharing}
\label{appd:3envs}
In Section~\ref{experiments}, we focus on the experiment setup with two environments: switching from Env-A to Env-B during training, and evaluate on both Env-A and Env-B during evaluation. To show the generality of the discussed problems and the proposed solutions, we include experiments with three environments in this section with both sequential training and parallel training. For each experiment setup, we include three different combinations of environment variations as listed in Table~\ref{env-table}.

\begin{table}[H]
\centering
\begin{tabular}{|c|c|c|c|}
\hline
               & \textbf{Env-Combination-1}                                                       & \textbf{Env-Combination-2}                                                       & \textbf{Env-Combination-3}                                                      \\ \hline
\textbf{Env-A} & Default                                                                          & Default                                                                          & Default                                                                         \\ \hline
\textbf{Env-B} & hip deformation = 0.2                                                            & hip deformation = 0.3                                                            & foot scale = 1.2                                                                \\ \hline
\textbf{Env-C} & \begin{tabular}[c]{@{}c@{}}hip deformation = 0.2\\ foot scale = 1.6\end{tabular} & \begin{tabular}[c]{@{}c@{}}hip deformation = 0.3\\ foot scale = 1.6\end{tabular} & \begin{tabular}[c]{@{}c@{}}foot scale = 1.2\\ ground texture = 0.1\end{tabular} \\ \hline
\end{tabular}
\caption{Combinations of different environment variations in the three environment experiments.}
\label{env-table}
\end{table}

\subsection{Forgetting and the effectiveness of Offline Distillation}
\begin{figure}[H]
\centering
\begin{subfigure}{\textwidth}
    \centering
    \includegraphics[width=0.5\linewidth]{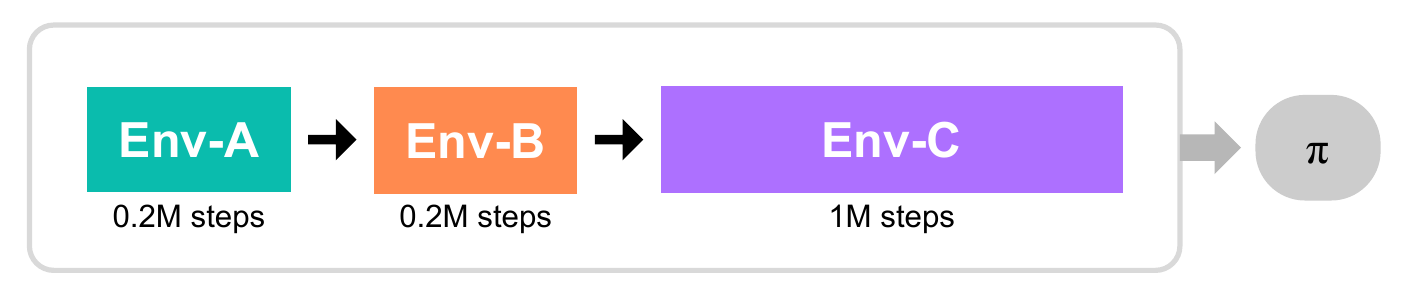}
    \caption{Experiment Setup: The agent experiences three stages across the online interaction phase.}
    \label{fig:appd_lifelong_scheme}
    \vspace{3mm}
\end{subfigure}
\begin{subfigure}{\textwidth}
    \centering
    \includegraphics[width=0.8\linewidth]{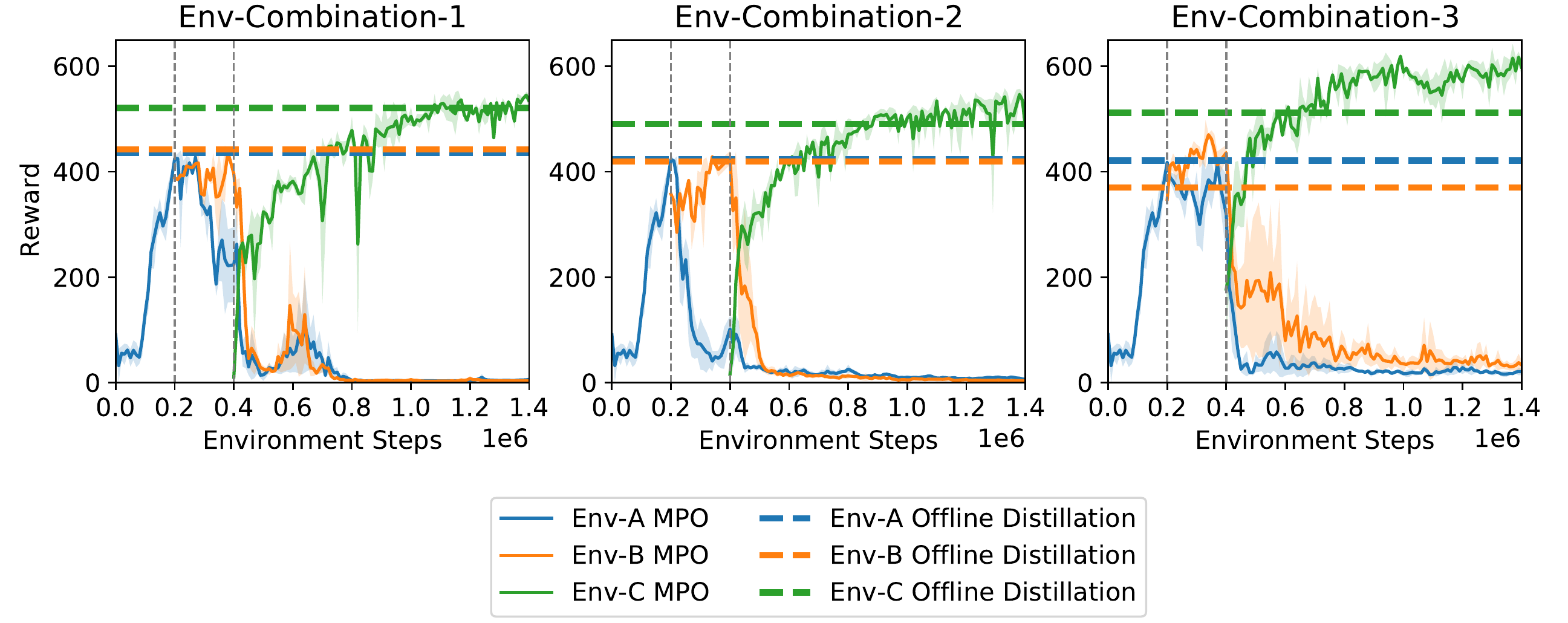}
    \caption{Reward curves}
    \label{fig:appd_lifelong_reward}
\end{subfigure}
\caption{Lifelong learning experiments with three stages. The policy trained with MPO during online interaction experiences significant forgetting on previous environments. With Offline Distillation at the end, the policy can recover the performance effectively over all the previous environments.}
\label{fig:appd_lifelong}
\end{figure}

In this section, we demonstrate the forgetting issue and the effectiveness of the Offline Distillation Pipeline with a three-stage experiment. As shown in Figure~\ref{fig:appd_lifelong_scheme}, the agent first collects data in Env-A for 0.2M steps, switches to Env-B for 0.2M steps, and eventually experiences Env-C for 1M steps. During this online interaction phase, the agent is trained with MPO and keeps all the history data in the replay buffer. We plot the evaluation performance for all three environments across this procedure in Figure~\ref{fig:appd_lifelong_reward}. Similar to the observations from Section~\ref{result:pipeline}, the agent experiences significant performance drop on the previous environments after the switches. At the end of the online interaction phase, we run Offline Distillation on the entire dataset indicated as the dotted lines in Figure~\ref{fig:appd_lifelong_reward}. With Offline Distillation, we can recover a policy that works on all the previous environments.

\subsection{Imbalance experience in offline distillation}
\begin{figure}[H]
\centering
\begin{subfigure}{\textwidth}
    \centering
    \includegraphics[width=0.5\linewidth]{figures/3envs-type3.pdf}
    \caption{Experiment Setup: The agent experiences three stages across the online interaction phase.}
    \label{fig:appd_type3}
\end{subfigure}
\begin{subfigure}{\textwidth}
    \centering
    \includegraphics[width=0.8\linewidth]{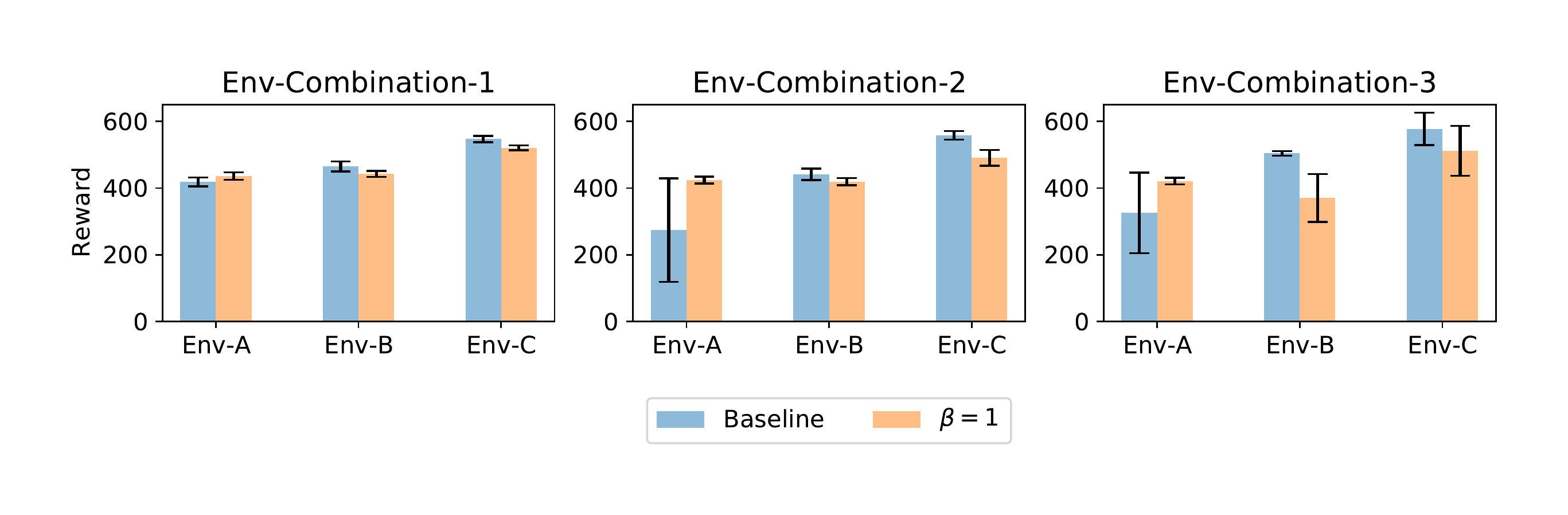}
    \vspace{-5mm}
    \caption{Final performance of Offline Distillation.}
    \label{fig:appd_imbalance_reward_type3}
\end{subfigure}
\caption{Evaluation of Offline Distillation with different transformation functions with a three-stage experiment setup.}
\label{fig:appd_imbalance_type3}
\end{figure}

We construct additional experiment setups to demonstrate the imbalance issue in Offline Distillation and verify the proposed solution of using a more conservative transformation function. First, we follow the setup from the previous section where the agent experiences three environments consecutively (Figure~\ref{fig:appd_type3}). We compare the results of Offline Distillation with an indicator function (Baseline) and an exponential function with $\beta=1$. From Figure~\ref{fig:appd_imbalance_reward_type3}, the baseline has a significant performance drop in Env-A for Env-Combination-2 and Env-Combination-3, similar to the results from Section~\ref{result:imbalance}. With the exponential function with $\beta=1$, CRR training during the distillation phase is more reliable across all the environments.

\begin{figure}[H]
\centering
\begin{subfigure}{\textwidth}
    \centering
    \includegraphics[width=0.4\linewidth]{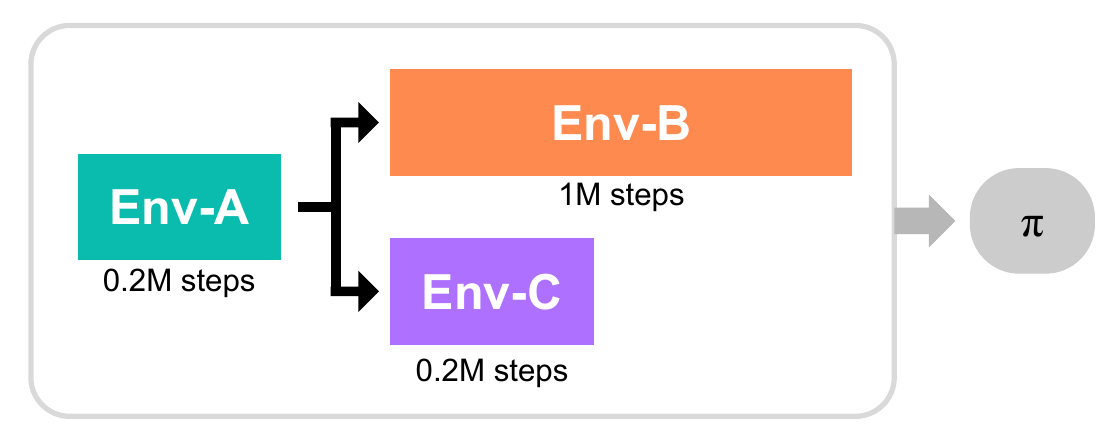}
    \caption{Experiment Setup: Parallel training during the online interaction phase (type-A).}
    \label{fig:appd_type2}
\end{subfigure}
\begin{subfigure}{\textwidth}
    \centering
    \includegraphics[width=0.8\linewidth]{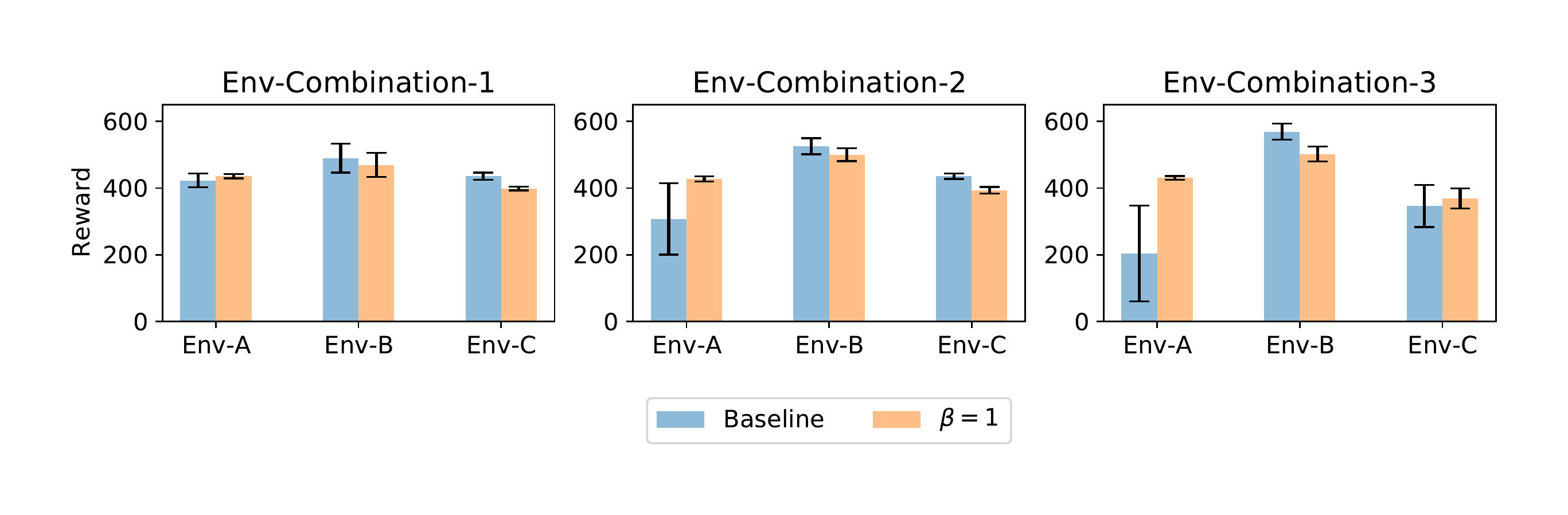}
    \vspace{-5mm}
    \caption{Final performance of Offline Distillation.}
    \label{fig:appd_imbalance_reward_type2}
\end{subfigure}
\caption{Evaluation of Offline Distillation with different transformation functions with a parallel training experiment setup with three environments (type-A).}
\label{fig:appd_imbalance_type2}
\end{figure}

\begin{figure}[H]
\centering
\begin{subfigure}{\textwidth}
    \centering
    \includegraphics[width=0.4\linewidth]{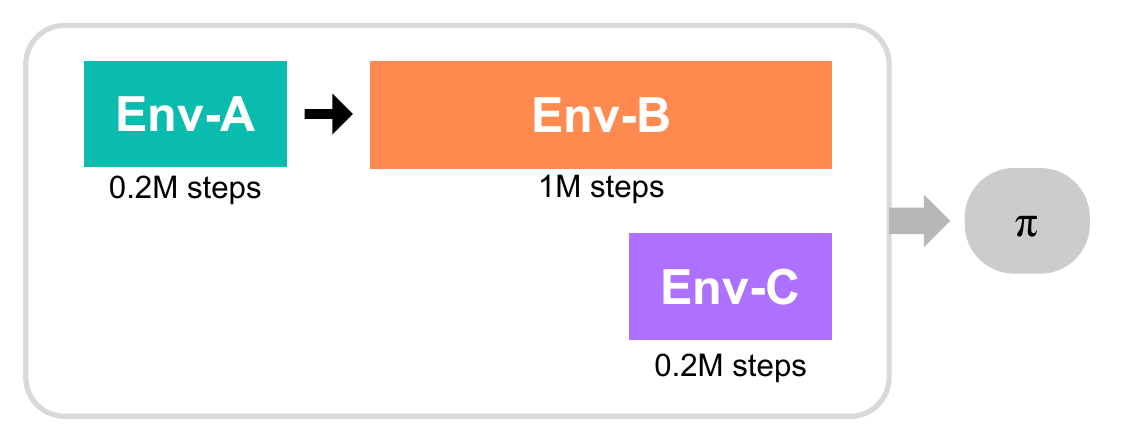}
    \caption{Experiment Setup: Parallel training during the online interaction phase (type-B).}
    \label{fig:appd_type1}
\end{subfigure}
\begin{subfigure}{\textwidth}
    \centering
    \includegraphics[width=0.8\linewidth]{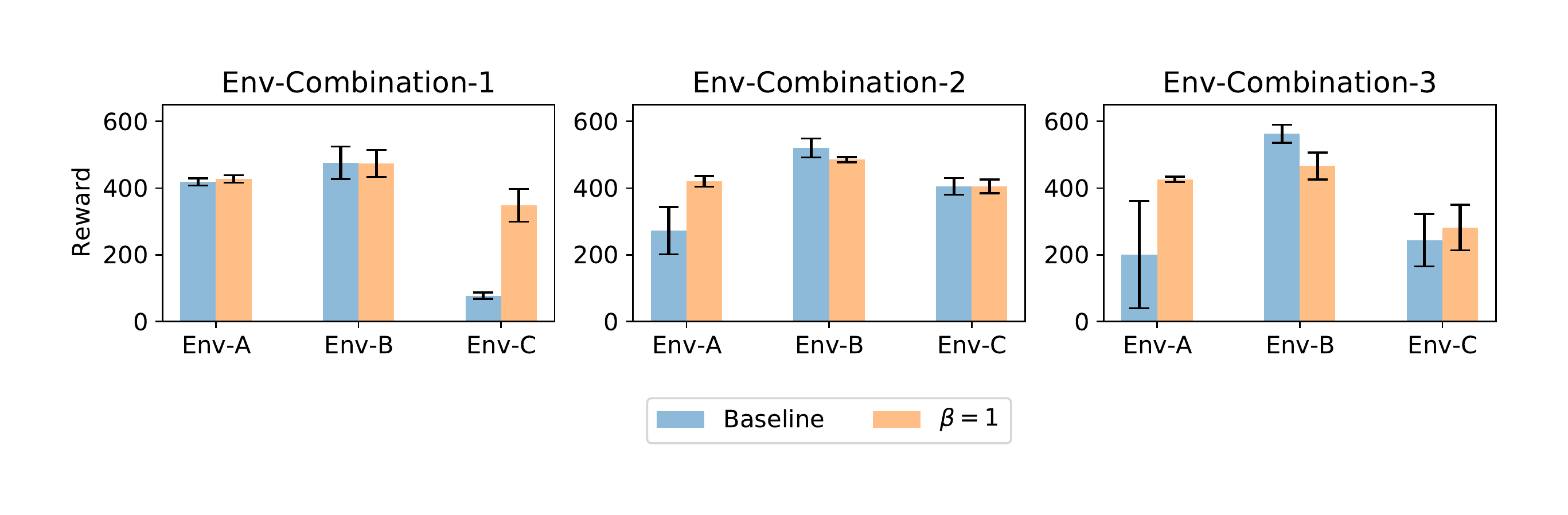}
    \vspace{-5mm}
    \caption{Final performance of Offline Distillation.}
    \label{fig:appd_imbalance_reward_type1}
\end{subfigure}
\caption{Evaluation of Offline Distillation with different transformation functions with a parallel training experiment setup with three environmentsv (type-B).}
\label{fig:appd_imbalance_type1}
\end{figure}

We evaluate two additional experiment setup with parallel training. In Figure~\ref{fig:appd_type2}, the agent is first trained on Env-A. After 0.2M steps, the agent is copied into two different agents (including the policy, the Q-function, and the replay buffer). Two copies of the agents continues the training on Env-B and Env-C independently. At the end of the online interaction phase, we run Offline Distillation on the entire dataset. In Figure~\ref{fig:appd_type1}, one agent is trained on Env-A and then switches to Env-B. Another agent is trained from scratch on Env-C. The offline distillation is performed on the combined dataset of these three environments. For both experiment setups, the baseline often experiences performance drop in Env-A. For the setup in Figure~\ref{fig:appd_imbalance_reward_type1}, Env-C might also suffer from a performance drop since the dataset is trained from scratch and it is relatively small. As we discussed in Section~\ref{imbalance_issue}, both size and quality contributes to the performance drop of imbalanced dataset. For both setups, CRR with exponential function with $\beta=1$ can largely fix this issue.

